\documentclass{article}

\usepackage{arxiv}

\usepackage[utf8]{inputenc} 
\usepackage[T1]{fontenc}    
\usepackage{hyperref}       
\usepackage{url}            
\usepackage{booktabs}       
\usepackage{amsfonts}       
\usepackage{nicefrac}       
\usepackage{microtype}      
\usepackage{graphicx}
\usepackage[square,numbers,sort&compress]{natbib}
\usepackage{doi}

\usepackage{float}
\usepackage{multirow}
\usepackage{amsmath,amssymb,amsthm}
\usepackage{bm}
\usepackage{braket}
\usepackage{mathrsfs}
\usepackage{algorithm}
\usepackage{algpseudocode}
\usepackage[framemethod=tikz]{mdframed}
\usepackage{tikz}
\usetikzlibrary{decorations.pathreplacing,arrows.meta}
\usepackage[normalem]{ulem}
\usepackage{cleveref}       

\graphicspath{{./figures/}}

\title{Spectral Path Regression: Directional Chebyshev Harmonics for Interpretable Tabular Learning}

\author{%
  Milo Coombs \\
  \texttt{milo.coombs7@gmail.com}
}

\renewcommand{\shorttitle}{Spectral Path Regression: Directional Chebyshev Harmonics for Interpretable Tabular Learning}

\hypersetup{
  pdftitle={Spectral Path Regression: Directional Chebyshev Harmonics for Interpretable Tabular Learning},
  pdfauthor={Milo Coombs},
  pdfkeywords={spectral methods, approximation theory, tabular regression, interpretability},
}

\begin{document}
\maketitle

\begin{abstract}
Classical approximation bases such as Chebyshev polynomials provide principled and interpretable representations, but their multivariate tensor-product constructions scale exponentially with dimension and impose axis-aligned structure that is poorly matched to real tabular data. We address this by replacing tensorised oscillations with \emph{directional harmonic modes} of the form $\cos(\mathbf{m}^{\top}\arccos(\mathbf{x}))$, which organise multivariate structure by direction in angular space rather than by coordinate index. This representation yields a discrete spectral regression model in which complexity is controlled by selecting a small number of structured frequency vectors (spectral paths), and training reduces to a single closed-form ridge solve with no iterative optimisation. Experiments on standard continuous-feature tabular regression benchmarks show that the resulting models achieve accuracy competitive with strong nonlinear baselines while remaining compact, computationally efficient, and explicitly interpretable through analytic expressions of learned feature interactions.
\end{abstract}

\keywords{Chebyshev Polynomials \and Spectral Methods \and Directional Harmonics \and Sparse Modeling \and Tabular Regression \and Interpretability}

\section{Introduction}
Many problems in science and engineering are ultimately problems of approximation: when governing laws are unknown, complex relationships must be represented by structured surrogate functions rather than closed-form equations \cite{trefethen2019approximation}. From this perspective, supervised learning can be viewed as the task of constructing accurate and generalisable approximations from a finite amount of data.

Modern machine learning achieves this at scale using highly expressive models trained by stochastic optimisation, such as neural networks and ensemble methods \cite{lecun2015deep}. While effective, these models typically learn implicit representations whose internal geometry is difficult to inspect, making it challenging to reason about multivariate structure, interactions, and global behaviour.

This work revisits approximation theory from a geometric standpoint and asks a foundational question: \emph{what kind of basis is well aligned with high-dimensional continuous tabular data?} Classical bases with strong one-dimensional properties often extend poorly to high dimensions. This is because classical multivariate bases are most commonly constructed from tensor products of one-dimensional bases; these scale exponentially with dimension and enforce axis-aligned structure that is often misaligned with real data \cite{friedman2008predictive}, although structured decompositions such as tensor trains can mitigate the worst scaling while retaining an axis-organised representation.

We develop a new basis, taking inspiration from Chebyshev polynomials because of their near-minimax behaviour and numerical stability in one dimension \cite{mason2002chebyshev}. This basis is defined in the associated angular coordinates and reorganises multivariate oscillations by \textit{direction} rather than by \textit{coordinate index}, directly addressing the scaling and axis-alignment issues of classical bases. Concretely, this new basis consists of integer-frequency modes of the form $\cos(\mathbf{m}^{\top}\arccos(\mathbf{x}))$, from which we select a sparse set of such modes — \emph{spectral paths} — directly from data.

This new basis yields a discrete spectral model for tabular regression with several desirable properties. Once a set of spectral paths is fixed, training reduces to a single closed-form ridge regression solve, requiring no iterative optimisation. Model complexity is controlled explicitly through sparsity and interaction order, and learned components correspond to interpretable multivariate feature interactions. We demonstrate on standard continuous-feature tabular benchmarks that this approach achieves predictive accuracy competitive with strong nonlinear baselines while remaining compact, computationally efficient, and analytically transparent.

\subsection*{Contributions}
The main contributions of this work are as follows:
\begin{enumerate}
    \item \textbf{A directional multivariate approximation basis.}  
    We introduce a non-axis-aligned basis for multivariate approximation inspired by Chebyshev geometry, in which oscillatory structure is organised by direction in angular space rather than by coordinate-wise degree. This reformulation preserves the underlying Chebyshev harmonic structure while avoiding the combinatorial and geometric limitations of tensor-product constructions.

    \item \textbf{A sparse, closed-form, and interpretable spectral model.}  
    We derive a regression model based on a small number of directional spectral components, for which training reduces to a single closed-form ridge solve. The resulting models admit explicit analytic expressions that expose low-order feature interactions.

    \item \textbf{A structured greedy selection procedure for spectral paths.}  
    We propose a learning algorithm that constructs the model by selecting a small set of structured frequency vectors (\emph{spectral paths}) via a forward greedy procedure with exact validation-based evaluation and streaming Gram updates.

    \item \textbf{Empirical validation on tabular regression benchmarks.}  
    We demonstrate that the proposed representation yields competitive predictive performance on standard continuous-feature tabular datasets while remaining compact, computationally efficient, and stable under hyperparameter variation.
\end{enumerate}
The remainder of the paper is organised as follows. Section~2 reviews related work and notation. Section~3 introduces the Chebyshev foundations underlying the angular representation. Section~4 develops the directional change of basis and formalises spectral paths. Section~5 presents the resulting regression model and learning procedure. Section~6 reports experimental results, and Section~7 concludes.

An implementation of the spectral path model and all experimental code is available at \url{https://github.com/MiloCoombs2002/spectral-paths}.
\section{Related Work \& Notation}
\label{sec:related-work}

\subsection{Related Work}

The present work lies at the intersection of classical approximation theory, spectral representations, and learning methods for tabular data. A wide range of modern approaches employ harmonic or polynomial features, kernel constructions, or ensemble-based interaction discovery. Superficial similarities in functional form, however, often mask substantial differences in motivation, structure, and interpretation. Our aim here is not to provide an exhaustive survey, but to situate the proposed model relative to several closely related lines of work and to clarify the specific sense in which it departs from them. In particular, we emphasise that the use of cosine features alone does not characterise the approach: the geometric origin of the representation, the angular transformation inherited from Chebyshev theory, and the structured organisation of multivariate interactions are central to the contribution \cite{trefethen2019approximation, molnar2020interpretable}.

\paragraph{Spectral feature methods and kernel approximations.}
A number of popular methods employ cosine features of dot products, most notably Random Fourier Features (RFFs) \cite{rahimi2007random}, which approximate shift-invariant kernels by mapping inputs to finite-dimensional feature vectors of the form $\cos(\bm{\omega}^\top \mathbf{x} + b)$ with randomly sampled frequencies. At a formal level, this resembles the present model in that both involve cosine evaluations. The similarity, however, is largely superficial. RFFs operate directly in the original input space and are motivated by kernel approximation: frequencies are sampled stochastically to approximate an implicit reproducing kernel Hilbert space.

By contrast, our representation arises from classical approximation theory rather than kernel methods \cite{bach2017breaking}. Inputs are first mapped to angular coordinates via $\bm{\theta}=\arccos(\mathbf{x})$, under which polynomial degree corresponds to harmonic frequency. Frequencies are integer-valued and structured, and multivariate interactions are organised explicitly by direction in angular space. While any explicit feature map induces an associated kernel, the conceptual starting point and resulting structure are different: rather than approximating a kernel implicitly, we define a finite, explicit approximation basis whose elements and coefficients admit direct analytic interpretation.

\paragraph{Tree ensembles and tabular learning.}
Tree-based ensemble methods \cite{breiman2001random, friedman2001greedy} are among the strongest general-purpose models for tabular data and serve as important empirical baselines. These methods construct piecewise-constant approximations through recursive, axis-aligned partitions of feature space, with interactions arising from combinations of splits. While individual trees can be interpreted through their hierarchical decision rules, large ensembles typically yield complex, highly partitioned representations that are difficult to analyse globally.

In contrast, the spectral path model represents the target function as a smooth superposition of global oscillatory components, each corresponding to a structured interaction among a small subset of features. The distinction is therefore representational rather than algorithmic: interactions are encoded spectrally rather than through combinatorial partitioning, yielding compact models with explicit analytic structure.

\paragraph{Harmonic moments}
Harmonic moment representations such as Fourier-Mellin moments, Zernike moments, and Chebyshev-Fourier moments have been widely studied in image analysis as rotation-invariant feature descriptors \cite{flusser2009moments}. These methods expand image intensities in orthogonal radial and angular bases defined on the unit disk. While superficially similar in their use of harmonic components, their goal is typically invariant image description rather than sparse approximation of multivariate functions. The present work instead applies directional harmonic features derived from Chebyshev angular coordinates to supervised learning problems in tabular data.

\paragraph{Interpretability.}
A large body of work addresses interpretability through post-hoc explanation techniques or by restricting model classes \cite{rudin2019stop}. In this work, interpretability follows directly from the representation itself. Each learned component corresponds to an explicit harmonic interaction with a well-defined support, frequency, and coefficient, yielding an analytic expression for the fitted function. Detailed examples of this structure and its interpretation are provided in Section~6.

\paragraph{Symbolic Regression and Analytic Discovery}
The spectral path model shares a primary objective with Symbolic Regression (SR): the discovery of explicit, parsimonious analytic expressions that describe the relationship between input features and targets \cite{cranmer2020discovering}. Traditional SR typically employs genetic programming or simulated annealing to search the unconstrained space of mathematical operators, a process that is often stochastically unstable, computationally expensive, and prone to ``expression bloat".

In contrast, our approach can be viewed as a form of structured symbolic regression where the functional form is restricted to directional harmonic modes in Chebyshev angular space. This restriction transforms the discovery process from an open-ended combinatorial search into a disciplined selection of spectral paths. While standard SR requires iterative optimization, the spectral path model yields explicit analytic expressions - such as the multivariate interaction formulas shown in Section~5 - through a single closed-form ridge solve. This provides the transparency and interpretability of symbolic models while maintaining the training stability and computational efficiency of linear algebra.

\subsection{Notation}
Individual samples are represented as feature vectors
\[
    \mathbf{x}^{(n)} = \big(x^{(n)}_1,\dots,x^{(n)}_D\big)^\top \in \mathbb{R}^D,
    \qquad n = 1,\dots,N,
\]
where $D$ denotes the number of features, and $N$ the number of samples. We also work in angular coordinates defined componentwise by the transformation
\[
    \bm{\theta}^{(n)} = \arccos\!\left(\mathbf{x}^{(n)}\right) = \left(\theta^{(n)}_1, \ldots, \theta^{(n)}_D\right)^\top.
\]
\section{Background}
\label{sec:background}
Many supervised learning problems can be viewed through the lens of approximation: given finitely many samples from an unknown function on a bounded domain, the goal is to construct a surrogate that generalises globally rather than merely interpolating locally. Classical approximation theory addresses this problem by expanding functions in structured bases whose elements encode assumptions about smoothness, global error distribution, and numerical stability.

\subsection{Chebyshev Polynomials}
Chebyshev polynomials, unlike local expansions such as Taylor series, or globally periodic bases such as Fourier series, provide near-minimax control of the maximum approximation error on bounded intervals without imposing artificial boundary conditions \cite{trefethen2019approximation}. Crucially for learning, they admit a simple angular representation in which polynomial structure becomes harmonic. This representation exposes the oscillatory organisation underlying global approximation and serves as the starting point for our construction.

Chebyshev polynomials of the first kind, denoted by $T_n(x)$ arise from the problem of minimax approximation: among all degree-$n$ polynomials with fixed leading coefficient, $T_n (x)$ minimises the maximum deviation on the interval $[-1,1]$ \cite{rivlin2020chebyshev}. This property yields near-optimal control of global approximation error and leads to excellent numerical stability.

Each polynomial, index by $n$, admits the closed-form expression
\[
    T_n(x) = \cos(n\arccos(x)),
\]
and a smooth function can be expanded in a Chebyshev series,
\[
    y(x) = \sum_{n=0}^{\infty} W_n \,T_n(x),
\]
with coefficients $W_n$ obtained by orthogonal projection. Low-frequency modes capture coarse structure, while higher-frequency modes refine the approximation, leading to rapid convergence for smooth functions.

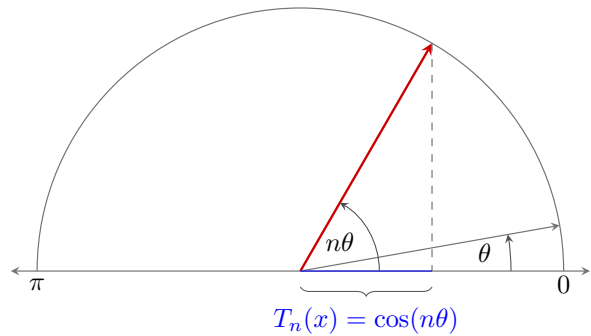
\begin{figure}[H]
\centering
\begin{tikzpicture}[scale=3.5,>=stealth]

\tikzset{
  scaff/.style={black!55, line width=0.45pt},
  proj/.style={black!55, dashed, line width=0.45pt},
  hero/.style={red!80!black, line width=0.9pt},
}

\draw[scaff,<->] (-1.1,0) -- (1.1,0);
\node[black] at (-1,-0.05) {$\pi$};
\node[black] at (1,-0.05) {$0$};
\draw[scaff] (-1,0) arc[start angle=180, end angle=0, radius=1];

\draw[thick, hero,-stealth] (0,0) -- ({cos(60)},{sin(60)});
\node[black] at (0.15,0.11) {$n\theta$};

\draw[thick, scaff,-stealth] (0,0) -- ({cos(10)},{sin(10)});
\node[black] at (0.7,0.07) {$\theta$};

\draw[proj] ({cos(60)},{sin(60)}) -- ({cos(60)},0);

\draw[blue] (0,0) -- ({cos(60)}, 0);

\draw[decorate, decoration={brace, mirror, amplitude=3.5pt}, black!65]
  (0, -0.06) -- ({cos(60)}, -0.06)
  node[blue, midway, below=4pt] {$T_n(x)=\cos(n\theta)$};

\draw[-stealth, black!70] (0.3,0) arc[start angle=0, end angle=60, radius=0.3];
\draw[-stealth, black!70] (0.8,0) arc[start angle=0, end angle=10, radius=0.8];

\end{tikzpicture}
\caption{Geometric interpretation of Chebyshev polynomials.
The input $x\in[-1,1]$ is lifted to an angle $\theta=\arccos(x)$ on the upper semicircle, rotated by a factor $n$, and projected onto the horizontal axis, yielding
$T_n(x)=\cos(n\theta)$.}
\label{fig:chebyshev-geometry1}
\end{figure}

As seen in Fig. \ref{fig:chebyshev-geometry1}, the basis function $T_n$ reveals a geometric interpretation; inputs $x \in [-1,1]$ are mapped to angles $\theta = \arccos(x)$ on the interval $[0,\pi]$. With this transformation, polynomial structure becomes harmonic: the index $n$ acts as a frequency that scales the angle, and the polynomial is obtained by projecting the resulting rotation through the cosine function.

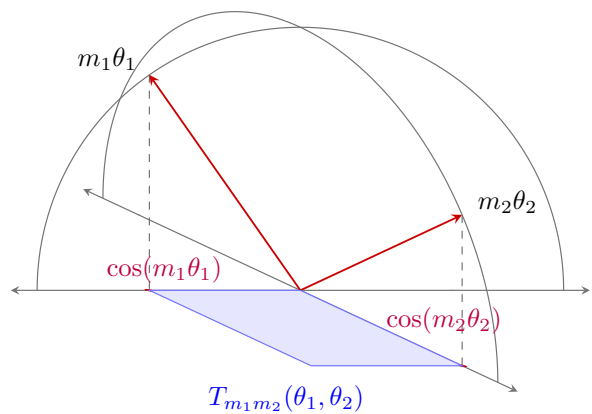
\begin{figure}[H]
\centering
\begin{tikzpicture}[scale=3.5,>=stealth]

\begin{scope}[
  x={(1cm,0cm)},
  y={(-0.75cm,0.35cm)},   
  z={(0cm,1cm)}
]

\def\R{1.0}        
\def\A{55}         
\def\B{35}         

\pgfmathsetmacro{\Ux}{cos(\A)}   
\pgfmathsetmacro{\Vy}{cos(\B)}   

\tikzset{
  scaff/.style={black!55, line width=0.45pt},
  proj/.style={black!55, dashed, line width=0.45pt},
  hero/.style={red!80!black, line width=0.7pt},
  fillrect/.style={fill=blue!10, draw=blue!55, line width=0.45pt}
}

\draw[scaff,<->] (-1.1,0,0) -- (1.1,0,0);   
\draw[scaff,<->] (0,-1.1,0) -- (0,1.1,0);   

\draw[scaff]
  plot[domain=180:0, samples=120]
  ({\R*cos(\x)}, 0, {\R*sin(\x)});

\draw[scaff]
  plot[domain=180:0, samples=120]
  (0, {\R*cos(\x)}, {\R*sin(\x)});

\draw[proj] ({\R*cos(180-\A)},0,{\R*sin(180-\A)}) -- ({\R*cos(180-\A)},0,0);
\fill[hero] ({\R*cos(180-\A)},0,0) circle (0.015);

\draw[hero,-stealth]
  (0,0,0) -- ({\R*cos(180-\A)}, 0, {\R*sin(180-\A)});
\node[black] at ({\R*cos(180-\A)-0.16}, 0, {\R*sin(180-\A)+0.06})
  {$m_1\theta_1$};

\draw[hero,-stealth]
  (0,0,0) -- (0, {\R*cos(180-\B)}, {\R*sin(180-\B)});
\node[black] at (0, {\R*cos(180-\B) -0.23}, {\R*sin(180-\B) +0.13})
  {$m_2\theta_2$};

\draw[proj] (0,{\R*cos(180-\B)},{\R*sin(180-\B)}) -- (0,{\R*cos(180-\B)},0);
\fill[hero] (0,{\R*cos(180-\B)},0) circle (0.015);

\draw[fillrect]
  (0,0,0)
  -- (-\Ux,0,0)
  -- (-\Ux,-\Vy,0)
  -- (0,-\Vy,0)
  -- cycle;

\draw[proj] (\Ux,0,0) -- (\Ux,0,0); 
\draw[proj] (0,\Vy,0) -- (0,\Vy,0);

\node[purple] at (-0.6*\Ux, 0.23, 0) {$\cos(m_1\theta_1)$};
\node[purple] at (0.30, -0.4*\Vy, 0) {$\cos(m_2\theta_2)$};

\node[blue] at (-1.7*\Ux, -1.5*\Vy, 0.02)
  {$T_{m_1 m_2}(\theta_1, \theta_2)$};

\end{scope}
\end{tikzpicture}
\caption{In a multivariate Chebyshev series, each variable is lifted to an angular coordinate, rotated independently by its frequency, and projected onto its coordinate axis. The resulting scalar projections combine multiplicatively, producing a rectangular “shadow” on the base plane whose area equals $\cos(m_1\theta_1)\times\cos(m_2\theta_2)$.}
\label{fig:chebyshev-geometry2}
\end{figure}

\subsection{Chebyshev Series and Multivariate Extensions}
Multivariate functions can also be approximated using Chebyshev polynomials using a basis constructed from tensor products of the one-dimensional basis.
\[
    y(x_1,\ldots,x_D) =
    \sum_{m_1=0}^{\infty}\!\cdots\!\sum_{m_D=0}^{\infty}
    W_{m_1,\ldots,m_D}
    \prod_{j=1}^D T_{m_j}(x_j).
\]
In this representation, each coordinate is rotated independently in its own angular dimension and projected back through the cosine function. As illustrated in Fig.~\ref{fig:chebyshev-geometry2}, multivariate structure arises through products of these univariate oscillations, with interactions encoded implicitly via combinations of frequencies across dimensions.

Truncating each index $m_j$ to order $M$ yields $(M+1)^D$ basis functions, leading to exponential growth in both storage and computation as  dimension $D$ increases \cite{novak2012tractability, bungartz2004sparse}. As a result, even moderate polynomial orders become infeasible in high-dimensional settings.

Beyond combinatorial scaling, the tensor-product construction also imposes a specific geometric organisation. Each angular coordinate $\theta_j$ is treated independently, and frequencies are assigned on a per-coordinate basis. Multivariate variation therefore emerges only through products of separable, axis-aligned oscillations. Capturing joint structure along oblique or low-dimensional directions in feature space typically requires assembling large collections of such terms, even when the underlying dependence is geometrically simple \cite{pinkus1999approximation}.

Crucially, once the angular coordinates are tensorised, the interpretation of $\bm{\theta}$ and $\mathbf{x}$ as a vector in a continuous angular space is no longer operative. The expansion does not admit a notion of direction, projection, or joint oscillation in $\bm{\theta}$; approximation proceeds by combining independent one-dimensional rotations rather than by modelling variation along directions in angular space. These limitations arise not from Chebyshev approximation itself, but from the tensor-product construction used to organise multivariate oscillations.

This observation motivates a reconsideration of how Chebyshev harmonic structure is indexed and combined in multiple dimensions. In the following section, we develop an alternative organisation that preserves the underlying angular representation while avoiding the combinatorial and geometric constraints imposed by tensorisation.
\section{From Tensor Products to Spectral Paths}
\label{sec:spectralpaths}
In this section we reinterpret multivariate Chebyshev approximation from a geometric and computational perspective. Rather than indexing oscillatory components by coordinate-wise degrees, we organise them by direction in angular space. This change of basis preserves the underlying Chebyshev harmonic structure while avoiding the combinatorial burden imposed by tensorisation. The resulting representation leads naturally to the notion of spectral paths.

\subsection{Directional Harmonics as a Change of Basis}

We consider basis functions of the form
\[
\phi_{\bm{m}}(\bm{\theta}) = \cos(\bm{m}^\top \bm{\theta}),
\qquad
\bm{m} \in \mathbb{Z}^D.
\]
Each integer vector $\bm{m}$ defines a global oscillation whose phase is given by the projection of $\bm{\theta}$ onto $\bm{m}$. The level sets of $\phi_{\bm{m}}$ are families of parallel hyperplanes in angular space, orthogonal to $\bm{m}$, so the function is constant along these hyperplanes and varies only in the direction of $\bm{m}$. In contrast to tensor-product constructions, oscillations therefore propagate along fixed directions rather than along individual coordinate axes.

This construction is best understood as a change of basis. It reorganises the same oscillatory content present in the tensor-product Chebyshev expansion, but indexed by direction rather than by coordinate-wise frequency tuples. In particular, the directional harmonic basis strictly contains the tensor-product Chebyshev basis: products of univariate cosines can always be written as finite sums of directional cosines. For example, in two dimensions,
\[
\underbrace{\cos(m_1\theta_1)\cos(m_2\theta_2)}_{T_{m_1 m_2}(\bm{\theta})}
=
\frac{1}{2}
\big[
\underbrace{\cos(m_1\theta_1 + m_2\theta_2)}_{\phi_{m_1 , m_2}(\bm{\theta})}
+
\underbrace{\cos(m_1\theta_1 - m_2\theta_2)}_{\phi_{m_1 , -m_2}(\bm{\theta})}
\big].
\]
More generally, any tensor-product Chebyshev polynomial can be expressed as a linear combination of directional harmonics with integer frequency vectors, as shown in the Appendix. As a result, the directional basis spans a superset of the classical Chebyshev space and inherits its approximation properties, including near-minimax control of uniform error. The reorganisation affects how oscillations are indexed and combined, not which functions can be represented.

Although the basis functions are cosine functions, expressivity is not limited by cosine’s even symmetry. The arguments of the cosines are integer combinations of angular coordinates $\theta_j = \arccos(x_j)$, so the composite functions $\cos(\bm{m}^\top \arccos(\bm{x}))$ are not even functions of the original input variables. For $m_j \geq 1$, each component $\cos(m_j \arccos(x_j))$ exhibits nonlinear and asymmetric dependence on $x_j$, and combining such terms across multiple coordinates yields a highly expressive hypothesis class capable of representing complex multivariate structure.

\begin{figure}[H]
\centering
\begin{tikzpicture}[scale=3.5,>=stealth]

\tikzset{
  scaff/.style={black!55, line width=0.45pt},
  proj/.style={black!55, dashed, line width=0.45pt},
  hero/.style={red!80!black, line width=0.9pt},
}

\draw[scaff,<->] (-1.1,0) -- (1.1,0);
\draw[scaff,<->] (0,-0.05) -- (0,1.1);
\node[black] at (0.08,1.1) {$\theta_2$};
\node[black] at (1.1,0.08) {$\theta_1$};

\draw[thick, hero,-stealth] (0,0) -- ({cos(60)},{sin(60)});
\node[hero] at ({cos(60)},{sin(60)+0.1}) {$(\theta_1^{(i)}, \theta_2^{(i)})$};

\draw[thick, scaff,-stealth] (0,0) -- ({cos(20)},{sin(20)});
\node[black] at ({cos(20)},{sin(20) + 0.1}) {$(m_1, m_2)$};

\draw[proj] ({cos(60)},{sin(60)}) -- ({cos(40)*cos(20)},{sin(20) * cos(40)});

\draw[blue] (0, 0) -- ({cos(40)*cos(20)},{sin(20) * cos(40)});

\draw[decorate, decoration={brace, mirror, amplitude=3.5pt}, black]
  (0.03, -0.03) -- ({cos(40)*cos(20) +0.03},{sin(20) * cos(40)-0.03})
  node[blue, midway, below=4pt, fill=white, inner sep=1pt, xshift=23pt] {$\phi_{\mathbf{m}}(\bm{\theta}^{(i)})/|\mathbf{m}|$};

\end{tikzpicture}
\caption{The new basis: directional harmonics defined by projections onto frequency vectors.}
\label{fig:chebyshev-geometry}
\end{figure}
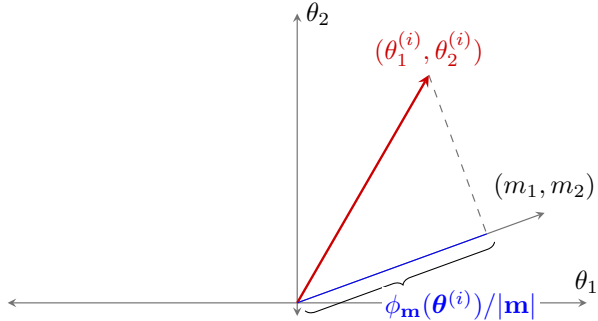

\subsection{Spectral Paths, Primitive Rays, and Computational Structure}
\paragraph{Computational structure.}
The directional basis aligns naturally with modern numerical computation. For a dataset with angular matrix $\Theta \in \mathbb{R}^{N \times D}$ and a collection of frequency vectors $\{\bm{m}_q\}_{q=1}^Q$, all directional phases can be computed simultaneously via a single matrix multiplication
\[
Z = \Theta M^\top,
\qquad\text{where}\qquad
(M)_{qj} = (m_q)_j.
\]
The corresponding features are obtained by applying the cosine function elementwise to $Z$. In contrast to tensor-product constructions, which require evaluating and multiplying separate univariate oscillations for each coordinate combination, the directional formulation reduces multivariate oscillations to dot products followed by pointwise nonlinearities. These operations are vectorisable, parallelisable, and well supported by modern hardware, with computational cost scaling linearly in both the number of features and the number of selected directions rather than exponentially in dimension.

\paragraph{Spectral paths and primitive rays.}
Each integer vector $\bm{m}\in\mathbb{Z}^D$ defines a single directional oscillation and is referred to as a \emph{spectral path}. These paths admit an internal structure: if $\bm{m} = r\,\bm{p}$ with $r\in\mathbb{Z}^+$ and $\bm{p}\in\mathbb{Z}^D$ primitive (i.e.\ $\bm{p}$ has no nontrivial integer divisor), then
\[
\cos(\bm{m}^\top\bm{\theta}) = \cos(r\,\bm{p}^\top\bm{\theta})
\]
represents the $r$th harmonic along the direction $\bm{p}$. Frequency vectors therefore organise naturally into \emph{primitive rays}, each supporting a ladder of harmonics. This representation factors out redundancy across harmonic multiples and groups oscillations by direction rather than by raw frequency index.

\paragraph{Sparsity and organisation.}
Classical Chebyshev approximation in one dimension is effective because spectral energy concentrates in low-order modes, with higher-order coefficients contributing progressively less for smooth functions. The same principle extends to multivariate settings: although the tensor-product basis obscures it combinatorially, meaningful structure is typically captured by a small number of low-order interactions. Restricting attention to $k$-sparse frequency vectors reflects this spectral decay intuition and yields interpretable models involving only a few interacting features.

Under this regime, most entries of the frequency matrix $M$ are zero, making dense representations inefficient. Grouping frequency vectors by primitive rays yields a sparse, graph-like organisation analogous to adjacency lists rather than adjacency matrices: only active harmonics along each direction are stored and evaluated. Viewed combinatorially, the support of $\bm{p}$ identifies the interacting features; viewed graph-theoretically, each spectral path corresponds to a hyperedge; and viewed computationally, primitive rays provide a structured mechanism for exploiting sparsity and harmonic structure simultaneously.

\subsection{Error Decay and Spectral Complexity}
\label{subsec:error-decay}

A key appeal of Chebyshev approximation in one dimension is its rapid reduction of \emph{global} approximation error as additional terms are included. This behaviour follows from the minimax property of Chebyshev polynomials and from their spectral interpretation via the identity $T_n(x)=\cos(n\arccos x)$. For analytic functions, Chebyshev coefficients decay exponentially with degree, so high accuracy is often achieved using only a small number of low-frequency modes.

In multivariate settings, however, error decay is inherently more nuanced. Classical tensor-product expansions admit exponential convergence only under strong and typically unrealistic isotropic smoothness assumptions, and there is no canonical notion of degree governing approximation quality. Instead, convergence depends critically on \emph{which} spectral components are retained, reflecting anisotropy, interaction structure, and directional variation in the target function.

The directional spectral path representation suggests a natural empirical hypothesis. By reorganising multivariate oscillations by direction in angular space rather than by coordinate-wise degree, and by selecting spectral paths greedily according to their contribution to predictive performance \cite{mallat1993matching, tropp2004greed}, the model implicitly prioritises low-frequency, high-energy components of the underlying function. In this respect, greedy path selection plays a role analogous to ordering coefficients by magnitude in a univariate Chebyshev expansion: early paths capture dominant coarse structure, while later additions refine the approximation.

\paragraph{Empirical hypothesis.}
Motivated by the exponential coefficient decay observed in one-dimensional Chebyshev series, we hypothesise that for sufficiently smooth targets whose dominant variation aligns with a small number of low-complexity directions, the prediction error of the spectral path model decreases rapidly as spectral paths are added, before stabilising once the principal structure has been captured. This behaviour should be understood as an empirical regularity rather than a formal convergence guarantee; the rate and extent of decay depend on noise, data alignment, and the particulars of the greedy selection strategy.

In the experiments that follow, we examine this hypothesis directly by measuring how prediction error evolves as spectral paths are incrementally added. Rather than asserting universal convergence behaviour, our goal is to demonstrate that across realistic tabular datasets the spectral path model exhibits rapid initial error reduction followed by stable saturation, consistent with the qualitative behaviour suggested by classical spectral approximation theory.
\section{The Model}
\label{sec:discrete-model}

The directional expansion developed in Section~\ref{sec:spectralpaths} defines a structured hypothesis class for multivariate approximation. We now turn this representation into a concrete regression model for continuous-feature tabular data. The key observation is that, once a finite set of spectral paths is fixed, learning reduces to ridge regression in an explicit feature space with a closed-form solution.

\subsection{Spectral Path Model}

Let $\bm{x}\in[-1,1]^D$ denote an input vector and define angular coordinates componentwise by
\[
\bm{\theta} = \arccos(\bm{x}) \in [0,\pi]^D.
\]
Given a finite set of spectral paths $\mathcal{M}=\{\bm{m}_1,\ldots,\bm{m}_Q\}\subset\mathbb{Z}^D$, we consider models of the form
\begin{equation}
\label{eq:spectral-path-model}
\hat y(\bm{\theta})
=
c_0
+
\sum_{q=1}^{Q}
A_q \cos(\bm{m}_q^\top \bm{\theta}),
\end{equation}
where $c_0\in\mathbb{R}$ is an intercept and $A_q\in\mathbb{R}$ are coefficients. The model is linear in its parameters and nonlinear in the inputs; all modelling choices are encoded in the selection of $\mathcal{M}$.

\subsection{Feature Map and Objective}

Given data $\{(\bm{x}^{(n)},y^{(n)})\}_{n=1}^N$, define spectral features
\[
\phi_q(\bm{\theta}) = \cos(\bm{m}_q^\top \bm{\theta}),
\qquad q=1,\ldots,Q,
\]
and let $\Phi\in\mathbb{R}^{N\times(1+Q)}$ denote the corresponding design matrix with an intercept column. For any fixed dictionary $\mathcal{M}$, coefficients are obtained by ridge regression,
\begin{equation}
\label{eq:ridge}
\bm{\beta}_\lambda
=
(\Phi^\top \Phi + \lambda I)^{-1}\Phi^\top \bm{y},
\end{equation}
where $\bm{\beta}=(c_0,A_1,\ldots,A_Q)^\top$ and $\lambda>0$ controls regularisation.

In practice, the design matrix need not be materialised explicitly. All computations can be expressed in terms of the Gram matrix $G=\Phi^\top\Phi$ and vector $b=\Phi^\top\bm{y}$, which can be accumulated in a streaming fashion. This yields memory usage independent of the number of samples and aligns naturally with large-scale tabular settings.

\subsection{Greedy Spectral Path Selection}
\label{subsec:greedy-selection}

The remaining task is to select a finite dictionary $\mathcal{M}$ from the infinite lattice $\mathbb{Z}^D$. We construct $\mathcal{M}$ using a forward greedy procedure that incrementally builds a sparse set of informative spectral paths.

At each iteration, the algorithm proposes candidate blocks of paths with small support sizes $k\in\mathcal{K}$, evaluates their contribution to validation performance, and appends the best-performing block to the active dictionary. Evaluation is performed exactly by solving the corresponding ridge system, rather than via residual-based screening.

\begin{algorithm}[t]
\caption{Greedy spectral path selection}
\label{alg:greedy-spectral-paths}
\begin{algorithmic}[1]
\Require Angular training data $(\bm{\theta}^{(n)},y^{(n)})$, validation data, sparsity levels $\mathcal{K}$, path budget $Q_{\max}$, block size $B$, ridge grid $\Lambda$
\State Initialise $\mathcal{M}\leftarrow \varnothing$
\State Initialise Gram terms $(G_{\mathrm{old}}, b_{\mathrm{old}})$
\State $\lambda_\star \leftarrow \texttt{None}$
\While{$|\mathcal{M}| < Q_{\max}$ and early stopping not triggered}
    \State Propose candidate blocks $\{\Delta\mathcal{M}_k\}_{k\in\mathcal{K}}$
    \For{each candidate block $\Delta\mathcal{M}_k$}
        \State Compute augmented normal equations via streaming updates
        \If{$\lambda_\star$ not fixed}
            \State Select best $\lambda\in\Lambda$ on validation set
        \Else
            \State Evaluate at $\lambda_\star$
        \EndIf
    \EndFor
    \State Select best block $\Delta\mathcal{M}^\star$
    \State Update $\mathcal{M}\leftarrow \mathcal{M}\cup\Delta\mathcal{M}^\star$
    \State Update Gram system
\EndWhile
\State \Return selected paths $\mathcal{M}$ and coefficients
\end{algorithmic}
\end{algorithm}

\paragraph{Structure of the search.}
The procedure restricts attention to sparse frequency vectors with small support sizes, so each selected path corresponds to a low-order interaction among a small subset of features. Candidates are explored in increasing order of complexity, ensuring that low-frequency structure is prioritised.

\paragraph{Exact evaluation.}
Unlike matching pursuit or residual-based methods, candidate blocks are evaluated by solving the corresponding ridge regression problem exactly (up to numerical precision). This ensures that selection is driven directly by validation performance rather than by surrogate criteria.

\paragraph{Computational strategy.}
All candidate evaluations are performed using streaming updates of the normal equations, avoiding explicit construction of the full design matrix. This keeps both memory usage and computational cost manageable even as the dictionary grows.

\paragraph{Regularisation and stopping.}
The regularisation parameter is selected on the validation set and fixed during the greedy search, with an optional final resweep after selection. The procedure terminates when validation performance saturates or a path budget is reached.

\paragraph{Summary.}
The resulting training procedure can be understood as a structured greedy search over sparse directional harmonics, combining:
\begin{itemize}
    \item low-order sparse interactions,
    \item exact validation-based selection, and
    \item efficient linear-algebraic updates.
\end{itemize}
Further implementation details are provided in the Appendix.
\section{Experiments}
\label{sec:experiments}

This section evaluates the spectral path model empirically. Rather than framing the experiments as a competition for state-of-the-art performance, our goal is to validate the central claims developed in Sections~\ref{sec:spectralpaths} and~\ref{sec:discrete-model}:

\begin{itemize}
\item Strong predictive accuracy can be achieved with a small number of structured spectral components.
\item Sparsity and compactness emerge naturally from the representation.
\item The learned models admit meaningful qualitative interpretation.
\item Training is stable and predictable under modest hyperparameter variation.
\end{itemize}

We report results on tabular regression benchmarks with predominantly continuous features.

\subsection{Experimental Setup}

All experiments use standard supervised regression benchmarks from the UCI repository \cite{Dua:2019}, OpenML \cite{Bischl2025OpenML}, and the Penn Machine Learning Benchmarks (PMLB) \cite{10.1093/bioinformatics/btab727}. Unless otherwise stated, datasets are split into training, validation, and test sets in a 60:20:20 ratio with a fixed random seed (42) for reproducibility.

Input features are transformed to lie in the bounded domain $[-1,1]$ prior to angular mapping. While several scaling strategies are possible, we employ a robust hyperbolic tangent transformation of the form
\[
x_j \;\mapsto\; \tanh\!\left(\frac{x_j - c_j}{s_j}\right),
\]
where $(c_j, s_j)$ are robust centring and scaling parameters estimated from the training data. In practice, we find that alternative scaling methods that map inputs to $[-1,1]$ yield qualitatively similar results; the choice of transformation primarily affects numerical conditioning rather than model structure.

The transformed inputs are then mapped to angular coordinates via $\theta = \arccos(x)$. Targets are centered on the training split and de-centered after prediction. Model selection and hyperparameter tuning are performed exclusively on the validation split.

For the spectral path model, ridge regularisation is used throughout. The regularisation parameter $\lambda$ is selected from a logarithmic grid on the validation split and held fixed for the final model. Spectral paths are selected greedily as described in Section~\ref{sec:discrete-model}. Candidate spectral paths are restricted to sparse frequency vectors with support sizes $k \in \{1,2,3,4\}$. The greedy procedure is allowed to select up to a maximum of $512$ paths, although in practice early stopping consistently terminates the search well before this limit. Ridge regularisation is tuned over a logarithmic grid $\lambda \in \{10^{-5}, \ldots, 10^{-1}\}$ on the validation split. Unless otherwise stated, these settings are used across all experiments.

Baseline models include:
\begin{itemize}
\item Ridge regression in the original input space, using standard $\ell_2$ regularisation with the regularisation parameter selected on the validation split,
\item Multilayer perceptrons (MLP), implemented as feedforward networks with two hidden layers of sizes 64 and 32 and ReLU activations, trained using standard optimisation procedures,
\item Gradient-boosted decision trees (XGBoost) \cite{chen2016xgboost}, using the default implementation with multi-threaded training.
\end{itemize}
All baselines are tuned using validation performance with reasonable default hyperparameter grids. Further implementation details are provided in the Appendix.

\subsection{Predictive Performance on Tabular Benchmarks}

We first assess predictive accuracy on a collection of standard tabular datasets. Table~\ref{tab:results-performance} reports test $R^2$ scores. Multilayer perceptrons and XGBoost are included as reference points representing widely used nonlinear learning methods. Reported values correspond to a representative run; performance across random splits is analysed separately.

\begin{table}[H]
\centering
\caption{Test $R^2$ performance across datasets. Best result per dataset in bold.}
\label{tab:results-performance}
\begin{tabular}{l l r r c c c c}
\toprule
Source & Dataset              & $N$   & $D$ & Spectral Paths & Ridge & MLP & XGBoost \\
\midrule
UCI    & Concrete Strength    & 1030  & 8   & 0.893 & 0.628 & 0.864 & \textbf{0.918} \\
UCI    & Energy (Heating)     & 768   & 8   & 0.998 & 0.907 & 0.995 & \textbf{0.998} \\
UCI    & Energy (Cooling)     & 768   & 8   & 0.979 & 0.888 & 0.981 & \textbf{0.992} \\
UCI    & Superconductivity    & 21263 & 81  & 0.838 & 0.737 & 0.891 & \textbf{0.902} \\
UCI    & Wine Quality         & 4898  & 11  & 0.352 & 0.259 & 0.353 & \textbf{0.423} \\
UCI    & Phishing Websites    & 11055 & 30  & 0.807 & 0.700 & \textbf{0.859} & 0.848 \\
OpenML & Concrete Slump       & 103   & 10  & \textbf{0.565} & 0.261 & 0.320 & 0.255 \\
OpenML & Yacht Hydrodynamics  & 308   & 7   & 0.985 & 0.568 & 0.989 & \textbf{0.998} \\
OpenML & Cancer Drug Response & 475   & 698 & \textbf{0.438} & 0.180 & 0.218 & 0.331 \\
OpenML & Aquatic Toxicity     & 546   & 9   & \textbf{0.488} & 0.475 & 0.393 & 0.341 \\
OpenML & Izmir Weather        & 1461  & 10  & \textbf{0.992} & 0.991 & 0.991 & 0.990 \\
OpenML & Ankara Weather       & 321   & 10  & \textbf{0.989} & 0.986 & 0.987 & 0.980 \\
PMLB   & Echocardiogram       & 17496 & 9   & \textbf{0.448} & 0.439 & 0.441 & 0.442 \\
PMLB   & Wind Speed           & 6574  & 14  & 0.797 & 0.778 & \textbf{0.800} & 0.778 \\
PMLB   & CPU Utilisation      & 8192  & 21  & \textbf{0.979} & 0.737 & 0.975 & 0.969 \\
\bottomrule
\end{tabular}
\end{table}
\begin{figure}[H]
  \centering
  \includegraphics[width=0.95\textwidth]{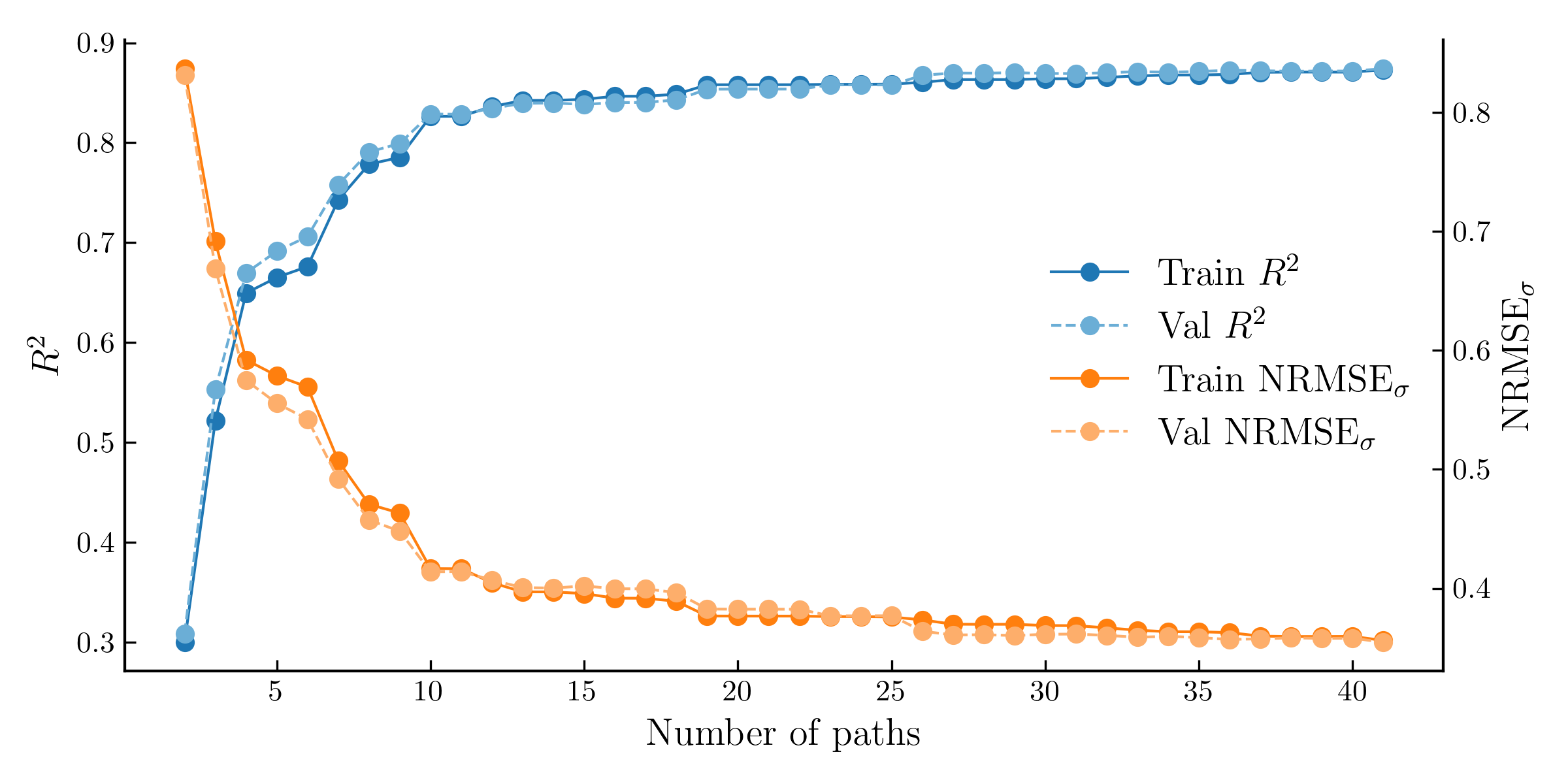}
  \caption{Training and validation performance of the discrete spectral model on the UCI Concrete dataset as a function of the number of paths $\mathcal{M}$. The model achieves stable generalisation after around $\mathcal{M}=9$, where validation $R^2$ plateaus and NRMSE$_\sigma$ reaches a minimum.}
  \label{fig:capacity-curves}
\end{figure}

Across datasets, the spectral path model achieves competitive performance relative to strong nonlinear baselines. While XGBoost or MLPs occasionally outperform it, the gap is modest given the substantially simpler training procedure and the explicit analytic structure of the learned model. These results establish that reorganising Chebyshev structure into directional harmonics does not sacrifice predictive accuracy on realistic tabular problems.

In addition to accuracy, the spectral path model exhibits competitive training and inference times relative to standard nonlinear baselines. Because training reduces to closed-form ridge regression once spectral paths are selected, runtime scales predictably with model size and remains stable across datasets. Detailed timing comparisons are reported in the Appendix.

\subsection{Model Compactness and Sparsity}
Predictive accuracy alone does not capture the primary advantage of the spectral path model. We therefore examine how performance evolves as a function of model size. Figure~\ref{fig:capacity-curves} shows training and validation performance on the Concrete Compressive Strength dataset as the number of selected spectral paths increases.

Performance improves rapidly with the addition of only a small number of paths and stabilises well before the dictionary becomes large. In particular, generalisation performance plateaus after approximately nine paths, corresponding to a tiny fraction of the combinatorial frequency space.

This behaviour contrasts sharply with tensor-product expansions or unstructured spectral methods, which typically require many more components to achieve comparable accuracy. The result supports the claim that tensorisation, rather than oscillatory approximation itself, is the true source of inefficiency in classical multivariate constructions.

\subsection{Qualitative Analysis of Learned Spectral Paths}
Beyond quantitative metrics, the spectral path model provides direct insight into the structure of the learned function.

\paragraph{Feature importance via analytic sensitivity.}
A key advantage of the spectral path model is that the fitted predictor admits an explicit analytic form (Equation~\ref{eq:spectral-path-model}). Unlike black-box models, this enables direct inspection of how the output varies with respect to each input feature.

We quantify feature importance using \emph{analytic sensitivity} with respect to the original (unscaled) input variables. Specifically, for each feature $x_j$, we define importance as the average absolute partial derivative of the fitted predictor,
\[
I_j
=
\frac{1}{N}
\sum_{n=1}^N
\left|
\frac{\partial \hat{y}(\mathbf{x}^{(n)})}{\partial x_j}
\right|.
\]
This quantity measures the expected magnitude of change in the prediction under infinitesimal perturbations of $x_j$, and therefore provides a natural notion of importance in the original feature space.

For the spectral path model,
\[
\hat{y}(\boldsymbol{\theta})
=
c_0 + \sum_{q=1}^Q A_q \cos(\mathbf{m}_q^\top \boldsymbol{\theta}),
\]
with $\theta_j = \arccos\!\big(\tanh((x_j - c_j)/s_j)\big)$, the partial derivatives can be computed exactly via the chain rule. In particular,
\[
\frac{\partial \hat{y}}{\partial x_j}
=
\frac{1}{s_j}
\operatorname{sech}\!\left(\frac{x_j - c_j}{s_j}\right)
\sum_{q=1}^Q
A_q\, m_{qj}\,
\sin\!\big(\mathbf{m}_q^\top \boldsymbol{\theta}\big),
\]
where $(c_j, s_j)$ are the fitted scaling parameters. This closed-form expression reflects the full structure of the model, incorporating coefficient magnitudes, harmonic order, and the nonlinear preprocessing.

\begin{figure}[t]
  \centering
  \includegraphics[width=\textwidth]{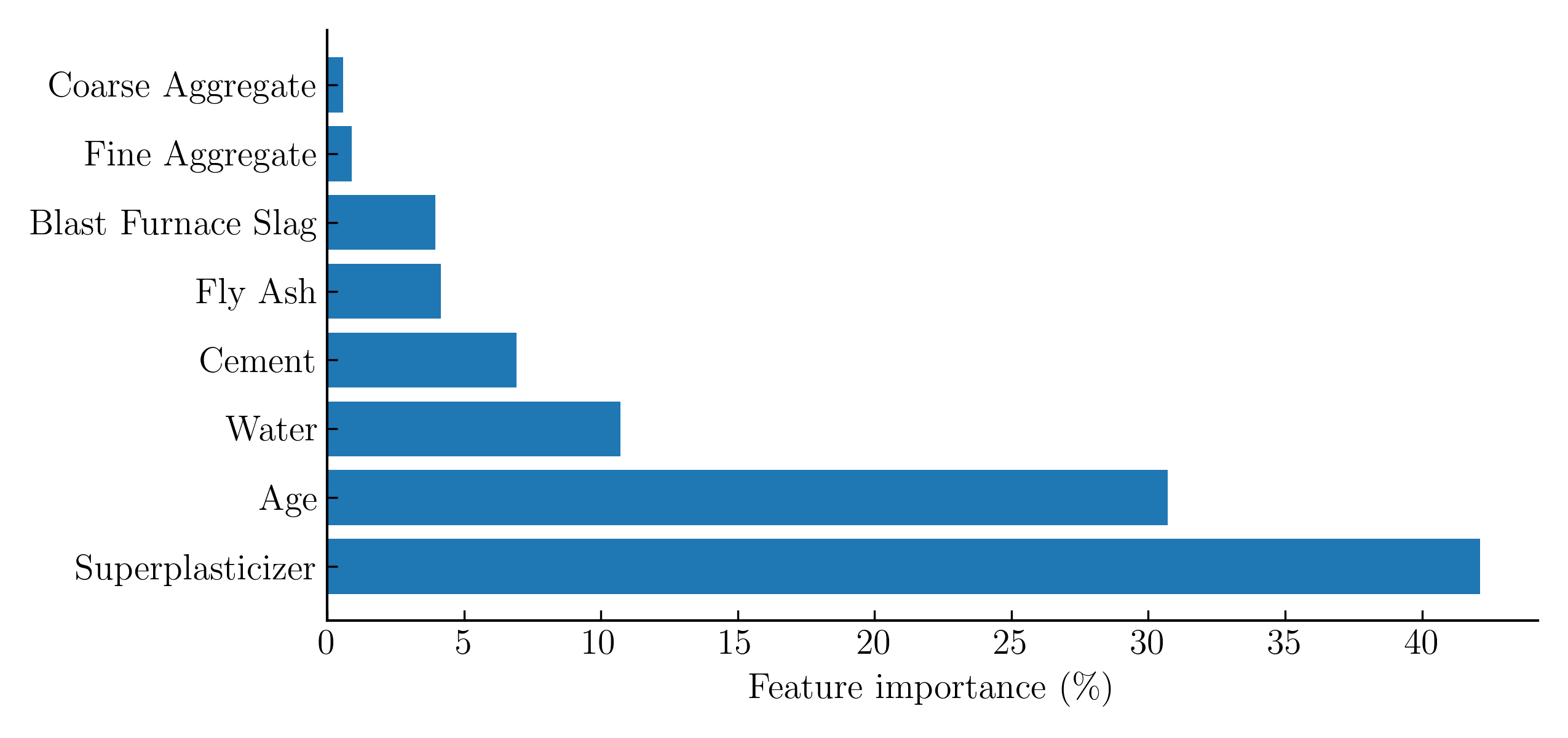}
  \caption{Feature importance on the Concrete dataset computed from normalised analytic sensitivities.}
  \label{fig:feature-importance}
\end{figure}

To facilitate comparison across features, the resulting importance scores are normalised to sum to one and reported as percentages. Figure~\ref{fig:feature-importance} shows these normalised sensitivity scores on the Concrete dataset. The learned importances align with domain intuition, highlighting Water and Blast Furnace Slag as the primary drivers of compressive strength while assigning relatively little weight to Age and Cement.

\paragraph{Discussion.}
This definition differs from coefficient-based or frequency-based measures in that it accounts for the entire computational pipeline, from input scaling through angular transformation to spectral expansion. As a result, it provides a faithful measure of feature influence in the original data domain. More broadly, the ability to compute such quantities exactly highlights a central advantage of the spectral path framework: interpretability arises directly from the analytic structure of the model, rather than from post-hoc approximation.

More strikingly, the fitted model yields an explicit analytic approximation of the target function. 
For the Concrete dataset, a truncated expansion takes the form
\begin{align}
\hat{y}(\mathbf{x})
\;\approx\;&
-8.15
+ 53.41\,\cos\!\big(\theta_{\text{Age}}\big)
+ 33.49\,\cos\!\big(\theta_{\text{Cement}}\big)
+ 31.99\,\cos\!\big(\theta_{\text{FlyAsh}}\big) \\
&- 29.32\,\cos\!\big(2\theta_{\text{Age}}\big)
+ 15.31\,\cos\!\big(3\theta_{\text{Age}}\big)
- 13.74\,\cos\!\big(2\theta_{\text{FlyAsh}}\big) \nonumber\\
&+ 11.39\,\cos\!\big(\theta_{\text{Slag}}\big)
+ 6.78\,\cos\!\big(2\theta_{\text{Slag}}\big)
- 6.72\,\cos\!\big(2\theta_{\text{SuperPlasticizer}}\big) \nonumber\\
&- 6.71\,\cos\!\big(\theta_{\text{Water}}\big)
+ 6.52\,\cos\!\big(3\theta_{\text{FlyAsh}}\big)
- 6.04\,\cos\!\big(4\theta_{\text{Age}}\big)
+ \cdots,
\end{align}
where each angular variable is defined by
\[
\theta_j
=
\arccos\!\left(
\tanh\!\left(\frac{x_j - c_j}{s_j}\right)
\right),
\]
with $(c_j, s_j)$ denoting the robust centring and scaling parameters learned from the training data.

\paragraph{Interpretation.}
Several structural patterns emerge immediately. The most influential feature is \textit{Superplasticizer}, which accounts for the largest proportion of the model's sensitivity. This indicates that small changes in superplasticizer dosage lead to substantial variations in predicted compressive strength, reflecting its well-known role in modifying workability and effective water–cement interaction.

\textit{Age} is the second most important feature, contributing a significant share of the overall sensitivity. This aligns with domain knowledge: curing time is a primary driver of strength development, and the presence of higher-order harmonic terms in the fitted model suggests strongly nonlinear maturation effects.

\textit{Water} and \textit{Cement} form a secondary tier of importance, indicating meaningful but less dominant contributions. Their influence is consistent with their central role in hydration chemistry, though the model attributes comparatively less marginal sensitivity to them than to superplasticizer and age.

In contrast, \textit{Fly Ash} and \textit{Blast Furnace Slag} exhibit moderate influence, while \textit{Fine Aggregate} and \textit{Coarse Aggregate} contribute only marginally. This suggests that, within this dataset, aggregate composition plays a relatively minor role in determining compressive strength compared to chemical and temporal factors.

Importantly, both the feature importance scores and the explicit functional form arise directly from the fitted model itself, without the need for auxiliary explanation methods. This stands in contrast to many modern machine learning models, where interpretation typically relies on post-hoc approximations. Here, interpretability is intrinsic: the model simultaneously provides a compact analytic expression of the target function and a principled measure of feature influence, both derived from the same underlying structure.
\subsection{Sensitivity and Stability}

Finally, we examine the stability of the spectral path model under variations in the ridge regularisation parameter $\lambda$.

The results show that test performance remains highly stable across several orders of magnitude in $\lambda$, with $R^2$ varying only within a narrow range. No sharply defined optimum is observed; instead, performance is broadly consistent across the entire range considered, indicating that generalisation is not strongly sensitive to the precise choice of regularisation strength.

At very small values of $\lambda$, performance exhibits slightly higher variability and occasional degradation, consistent with mild overfitting due to insufficient regularisation of higher-frequency or weakly supported components. For larger $\lambda$, performance remains competitive without a clear monotonic trend, suggesting that increasing regularisation does not significantly impair the model’s capacity.

Overall, the relatively flat performance profile indicates that ridge regularisation plays a secondary, stabilising role, while the dominant inductive bias is determined by the structure of the selected spectral paths.

\begin{figure}[H]
  \centering
  \includegraphics[width=0.9\textwidth]{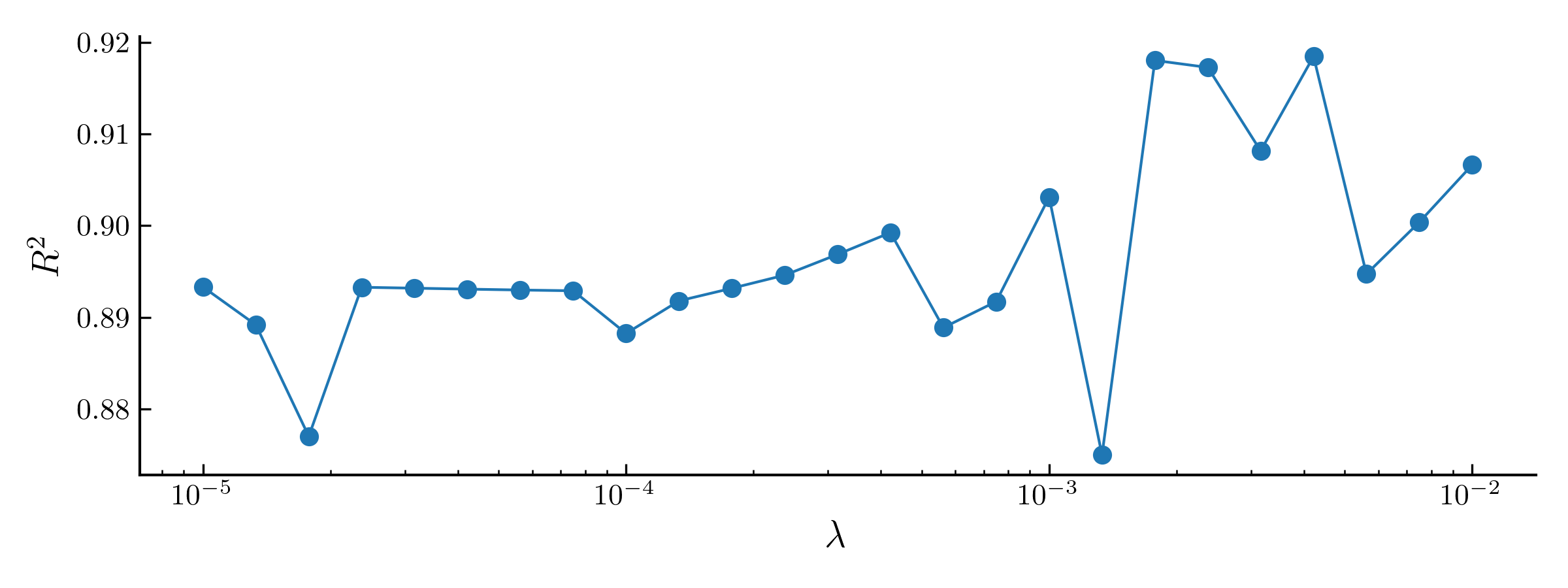}
  \caption{Test performance as a function of the ridge regularisation parameter $\lambda$ for the spectral path model on the UCI Concrete Compressive Strength dataset. The horizontal axis shows $\lambda$ on a logarithmic scale, while the vertical axis reports test $R^2$. Performance remains relatively stable across several orders of magnitude in $\lambda$, indicating limited sensitivity to precise regularisation strength and a broad region of near-optimal performance.}
  \label{fig:r2lambda}
\end{figure}

\begin{figure}[H]
  \centering
  \includegraphics[width=0.9\textwidth]{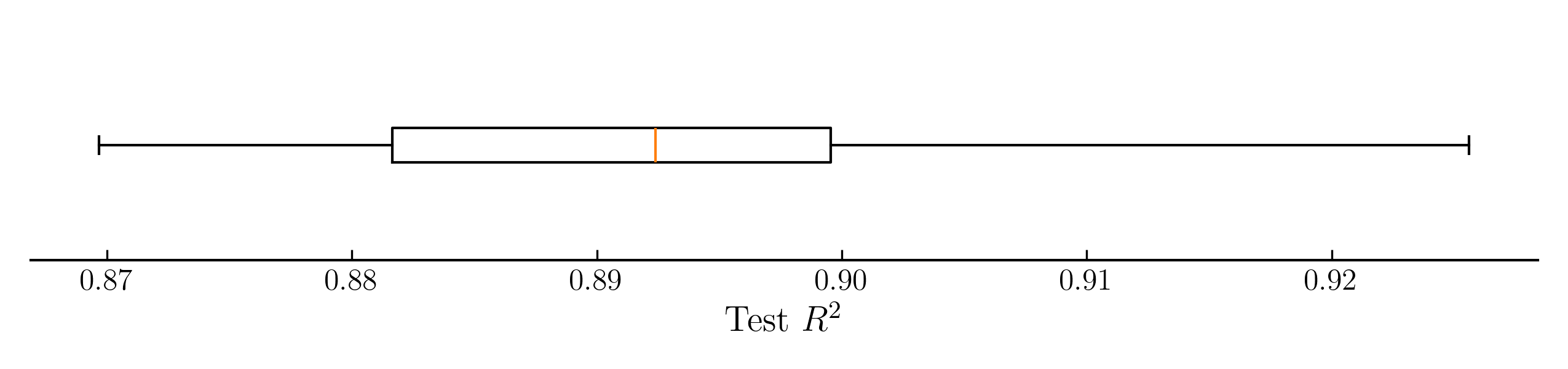}
  \caption{Box plot of test $R^2$ scores over 10 random train-validation-test splits. The spectral path model shows consistent performance across splits, with no extreme outliers, suggesting robustness to data partitioning and greedy path selection.}
  \label{fig:boxplot}
\end{figure}

Across all datasets examined, performance varies smoothly with $\lambda$ and exhibits low variance across splits. This suggests that the greedy path selection procedure is robust and that the learned models do not depend sensitively on particular hyperparameter choices.

\subsection{Summary}

Taken together, these experiments demonstrate that the spectral path model achieves strong predictive performance while using compact, structured representations. Its advantages lie not only in accuracy, but in sparsity, interpretability, and stability. The results support the central thesis of this work: that reorganising multivariate Chebyshev structure by direction rather than by tensor products yields models that are both effective and transparent.

\subsection{Further Work}

The present work focuses on scalar regression for tabular data, where the target can be written as a function of finitely many variables,
\[
    y(\theta_1,\ldots,\theta_D),
\]
and where a Chebyshev-inspired spectral representation yields a compact and interpretable hypothesis class. While this setting allows the core ideas to be developed cleanly, it also suggests several natural directions for extension.

\paragraph{Algorithmic and structural extensions.}
A first class of extensions concerns the construction and organisation of the spectral dictionary itself. In the current formulation, spectral paths are selected using a greedy forward procedure guided by validation performance. While effective in practice, alternative strategies may offer improved efficiency, robustness, or theoretical insight. These include residual-based criteria analogous to matching-pursuit methods, stability selection across resampled datasets, and group-wise selection at the level of primitive rays rather than individual harmonics. Similarly, regularisation could be refined to exploit internal spectral structure by penalising higher harmonics or larger interaction supports more strongly, reflecting prior beliefs about spectral decay and interaction order. Together, such developments would preserve the finite-dimensional and interpretable nature of the model while providing greater control over sparsity, stability, and inductive bias.

\paragraph{Broadening the learning scope.}
A second direction concerns extending the framework beyond scalar regression. The spectral path representation applies directly to classification by replacing the squared-loss objective with logistic or softmax likelihoods. Although closed-form solutions are then no longer available, the resulting optimisation problems remain convex and well behaved. Multi-output regression provides another natural extension: a shared set of spectral paths could be learned jointly across outputs, with output-specific coefficients capturing task-dependent structure. More broadly, the learned spectral paths may be viewed as a structured, data-adapted feature map that can be supplied to downstream learners, including linear models, tree ensembles, or neural networks. From this perspective, the spectral path model functions not only as a standalone predictor, but also as a principled feature-engineering mechanism grounded in approximation theory.

\paragraph{Beyond functions: functionals, operators, and scientific learning.}
Finally, the present theory addresses settings in which each input is a finite-dimensional vector. In many scientific and spatiotemporal problems, however, inputs are more naturally modelled as functions defined over a domain, and predictions take the form of functionals,
\[
    y \;=\; \mathcal{F}\!\big[X(\mathbf{u})\big],
\]
or operators,
\[
    Y(\mathbf{u}) \;=\; \mathcal{T}\!\big[X(\mathbf{u})\big].
\]
Extending the spectral path principle to these settings would require developing an approximation theory for functionals and operators that plays a role analogous to Chebyshev theory in the scalar case. Such extensions would align the present framework with modern operator-learning approaches in scientific machine learning, while preserving the emphasis on explicit structure, interpretability, and spectral organisation. We view the present work as establishing the finite-dimensional foundation upon which these broader generalisations may be built.

\paragraph{Hybrid spectral-conditional models.}
The spectral path model is well suited to targets that behave as smooth functions of their inputs, where global approximation and low-frequency structure dominate. However, many real-world tabular problems involve intrinsically conditional behaviour-threshold effects, regime switches, or logic that is naturally expressed through if-then rules or categorical splits. In such settings, tree-based models excel by representing functions as compositions of piecewise-constant regions, but they typically approximate smooth structure indirectly through large ensembles, leading to models that are difficult to interpret geometrically.

A natural direction for future work is therefore the development of hybrid models that combine conditional structure with spectral approximation. One possibility is a tree-like architecture in which internal nodes perform information-theoretic or variance-reducing splits, while leaf nodes contain low-complexity spectral path models that capture smooth behaviour within each region. Such a construction would allow conditional logic to be represented explicitly, while preserving the interpretability and approximation efficiency of directional Chebyshev expansions where smooth structure is present.

\section{Conclusion}
\label{sec:conclusion}

This paper revisits multivariate approximation from a geometric perspective and argues that the principal obstacle to scaling classical spectral methods is not oscillatory representation itself, but the axis-aligned tensor-product structure traditionally used to organise it. By working explicitly in Chebyshev angular coordinates and replacing tensorised oscillations with directional harmonic modes, we derive a compact spectral representation that captures multivariate interactions directly along directions in feature space.

From this representation we construct a discrete spectral regression model with a closed-form solution. Model complexity is controlled through the selection of a small number of structured frequency vectors, yielding sparse, interpretable expansions whose terms correspond to low-order feature interactions. Empirical results on standard tabular benchmarks demonstrate that these models achieve predictive performance competitive with strong nonlinear baselines while requiring no iterative optimisation and substantially lower training cost. Importantly, the learned models admit explicit analytic expressions, enabling direct inspection of both global structure and interaction effects.

Beyond empirical performance, the primary contribution of this work lies in reframing how classical approximation theory can be adapted to modern learning problems. By aligning representation with the geometry of data rather than with coordinate axes, the proposed framework reconciles several traditionally competing objectives: scalability, interpretability, and principled approximation. The results suggest that many of the limitations associated with multivariate polynomial and spectral methods arise from representational choices rather than from fundamental theoretical constraints.

We view the spectral path formulation as a foundation rather than a final model. Numerous extensions are possible, including improved path selection strategies, alternative regularisation schemes, classification and multi-output variants, and connections to structured feature engineering and interaction discovery. More broadly, the ideas developed here point toward a systematic programme for extending approximation-theoretic principles beyond scalar functions to functionals and operators, with potential applications in scientific machine learning and beyond.

Taken together, this work demonstrates that revisiting classical theory through a geometric lens can yield learning models that are both effective and transparent. By reorganising multivariate approximation around directional structure, we hope to encourage further exploration of interpretable, theory-driven alternatives to black-box models in high-dimensional settings.

\bibliographystyle{unsrtnat}
\bibliography{references}

@book{trefethen2019approximation,
  title={Approximation theory and approximation practice, extended edition},
  author={Trefethen, Lloyd N},
  year={2019},
  publisher={SIAM}
}

@article{lecun2015deep,
  title={Deep learning},
  author={LeCun, Yann and Bengio, Yoshua and Hinton, Geoffrey},
  journal={nature},
  volume={521},
  number={7553},
  pages={436--444},
  year={2015},
  publisher={Nature Publishing Group UK London}
}

@article{friedman2008predictive,
  title={Predictive learning via rule ensembles},
  author={Friedman, Jerome H and Popescu, Bogdan E},
  year={2008}
}

@book{mason2002chebyshev,
  title={Chebyshev polynomials},
  author={Mason, John C and Handscomb, David C},
  year={2002},
  publisher={Chapman and Hall/CRC}
}

@book{molnar2020interpretable,
  title={Interpretable machine learning},
  author={Molnar, Christoph},
  year={2020},
  publisher={Lulu. com}
}

@article{rahimi2007random,
  title={Random features for large-scale kernel machines},
  author={Rahimi, Ali and Recht, Benjamin},
  journal={Advances in neural information processing systems},
  volume={20},
  year={2007}
}

@article{bach2017breaking,
  title={Breaking the curse of dimensionality with convex neural networks},
  author={Bach, Francis},
  journal={Journal of Machine Learning Research},
  volume={18},
  number={19},
  pages={1--53},
  year={2017}
}

@article{friedman2001greedy,
  title={Greedy function approximation: a gradient boosting machine},
  author={Friedman, Jerome H},
  journal={Annals of statistics},
  pages={1189--1232},
  year={2001},
  publisher={JSTOR}
}

@article{breiman2001random,
  title={Random forests},
  author={Breiman, Leo},
  journal={Machine learning},
  volume={45},
  number={1},
  pages={5--32},
  year={2001},
  publisher={Springer}
}

@book{flusser2009moments,
  title={Moments and moment invariants in pattern recognition},
  author={Flusser, Jan and Zitova, Barbara and Suk, Tomas},
  year={2009},
  publisher={John Wiley \& Sons}
}

@article{rudin2019stop,
  title={Stop explaining black box machine learning models for high stakes decisions and use interpretable models instead},
  author={Rudin, Cynthia},
  journal={Nature machine intelligence},
  volume={1},
  number={5},
  pages={206--215},
  year={2019},
  publisher={Nature Publishing Group UK London}
}

@book{rivlin2020chebyshev,
  title={Chebyshev polynomials},
  author={Rivlin, Theodore J},
  year={2020},
  publisher={Courier Dover Publications}
}

@article{cranmer2020discovering,
  title={Discovering symbolic models from deep learning with inductive biases},
  author={Cranmer, Miles and Sanchez Gonzalez, Alvaro and Battaglia, Peter and Xu, Rui and Cranmer, Kyle and Spergel, David and Ho, Shirley},
  journal={Advances in neural information processing systems},
  volume={33},
  pages={17429--17442},
  year={2020}
}

@article{bungartz2004sparse,
  title={Sparse grids},
  author={Bungartz, Hans-Joachim and Griebel, Michael},
  journal={Acta numerica},
  volume={13},
  pages={147--269},
  year={2004},
  publisher={Cambridge University Press}
}

@book{novak2012tractability,
  title={Tractability of Multivariate Problems: Volume III: Standard Information for Operators},
  author={Novak, Erich and Wo{\'z}niakowski, Henryk},
  volume={18},
  year={2012},
  publisher={European Mathematical Society Publishing House}
}

@article{pinkus1999approximation,
  title={Approximation theory of the MLP model in neural networks},
  author={Pinkus, Allan},
  journal={Acta numerica},
  volume={8},
  pages={143--195},
  year={1999},
  publisher={Cambridge University Press}
}

@article{mallat1993matching,
  title={Matching pursuits with time-frequency dictionaries},
  author={Mallat, St{\'e}phane G and Zhang, Zhifeng},
  journal={IEEE Transactions on signal processing},
  volume={41},
  number={12},
  pages={3397--3415},
  year={1993},
  publisher={IEEE}
}

@article{tropp2004greed,
  title={Greed is good: Algorithmic results for sparse approximation},
  author={Tropp, Joel A},
  journal={IEEE Transactions on Information theory},
  volume={50},
  number={10},
  pages={2231--2242},
  year={2004},
  publisher={IEEE}
}

@misc{Dua:2019,
  author = {Dua, Dheeru and Graff, Casey},
  institution = {University of California, Irvine, School of Information and Computer Sciences},
  title = {{UCI} Machine Learning Repository},
  url = {http://archive.ics.uci.edu/ml},
  year = 2017
}

@article{Bischl2025OpenML,
  author = {Bischl, Bernd and Casalicchio, Giuseppe and Das, Taniya and Feurer, Matthias and Fischer, Sebastian and Gijsbers, Pieter and Mukherjee, Subhaditya and Müller, Andreas C. and Németh, László and Oala, Luis and Purucker, Lennart and Ravi, Sahithya and van Rijn, Jan N. and Singh, Prabhant and Vanschoren, Joaquin and van der Velde, Jos and Wever, Marcel},
  title = {OpenML: Insights from 10 years and more than a thousand papers},
  journal = {Patterns},
  year = {2025},
  volume = {6},
  number = {7},
  month = {07},
  doi = {10.1016/j.patter.2025.101317},
  url = {https://doi.org/10.1016/j.patter.2025.101317},
  publisher = {Elsevier},
  issn = {2666-3899}
}

@article{10.1093/bioinformatics/btab727,
    author = {Romano, Joseph D and Le, Trang T and La Cava, William and Gregg, John T and Goldberg, Daniel J and Chakraborty, Praneel and Ray, Natasha L and Himmelstein, Daniel and Fu, Weixuan and Moore, Jason H},
    title = {PMLB v1.0: an open-source dataset collection for benchmarking machine learning methods},
    journal = {Bioinformatics},
    volume = {38},
    number = {3},
    pages = {878-880},
    year = {2021},
    month = {10},
    issn = {1367-4803},
    doi = {10.1093/bioinformatics/btab727},
    url = {https://doi.org/10.1093/bioinformatics/btab727},
    eprint = {https://academic.oup.com/bioinformatics/article-pdf/38/3/878/49007845/btab727.pdf},
}

@article{chen2016xgboost,
  title={XGBoost: A Scalable Tree Boosting System},
  author={Chen, Tianqi},
  journal={Cornell University},
  year={2016}
}
\newpage
\appendix
\section{Directional Harmonics as a Superset of Tensor-Product Chebyshev Bases}
\label{app:superset-proof}

This appendix clarifies the relationship between the directional harmonic basis introduced in the main text and the classical tensor-product Chebyshev basis. In particular, we show that tensor-product Chebyshev basis functions lie within the linear span of directional cosine features, while emphasising that the resulting learning model is strictly more expressive due to the relaxation of coefficient-tying constraints. We also discuss the implications of this relationship for approximation guarantees.

\subsection{Span Inclusion in Two Dimensions}

We begin with the two-dimensional case, where the relationship can be seen most transparently. Consider a tensor-product Chebyshev basis function expressed in angular coordinates,
\[
\cos(m_1 \theta_1)\cos(m_2 \theta_2),
\qquad m_1,m_2\in\mathbb{Z}^+.
\]
Using the standard trigonometric identity,
\[
\cos a \cos b
=
\frac{1}{2}\big[\cos(a+b)+\cos(a-b)\big],
\]
we obtain
\[
\cos(m_1 \theta_1)\cos(m_2 \theta_2)
=
\frac{1}{2}\cos(m_1\theta_1 + m_2\theta_2)
+
\frac{1}{2}\cos(m_1\theta_1 - m_2\theta_2).
\]
Each term on the right-hand side is a directional harmonic of the form
\[
\cos(\bm{m}^\top \bm{\theta}),
\qquad
\bm{m}\in\mathbb{Z}^2,
\]
with frequency vectors $(m_1,m_2)$ and $(m_1,-m_2)$ respectively. Thus, every two-dimensional tensor-product Chebyshev basis element can be expressed as a finite linear combination of directional cosine features. This establishes that the classical tensor-product basis is contained within the span of the directional basis in two dimensions.

\subsection{Extension to Higher Dimensions}

The same argument extends directly to higher dimensions. A $D$-dimensional tensor-product cosine term takes the form
\[
\prod_{j=1}^D \cos(m_j \theta_j).
\]
Repeated application of the identity for products of cosines yields an expansion as a finite sum of directional harmonics,
\[
\prod_{j=1}^D \cos(m_j \theta_j)
=
2^{-(D-1)}
\sum_{\substack{\boldsymbol{\sigma}\in\{\pm1\}^D \\ \sigma_1=1}}
\cos\!\left(
\sum_{j=1}^D \sigma_j m_j \theta_j
\right),
\]
where the sum ranges over all sign patterns with the first sign fixed to avoid duplication. Each term corresponds to a directional cosine with integer frequency vector
\[
\bm{m}_{\boldsymbol{\sigma}}
=
(\sigma_1 m_1,\ldots,\sigma_D m_D)\in\mathbb{Z}^D.
\]
Consequently, every tensor-product Chebyshev basis function can be written as a linear combination of directional harmonics. The directional basis therefore spans a superset of the classical tensor-product Chebyshev space.

\subsection{Coefficient Tying and Model Expressivity}

While the span inclusion result establishes representational containment, it is important to distinguish this from equivalence of models. In the tensor-product expansion, each basis element corresponds to a specific linear combination of directional harmonics with \emph{tied coefficients}. For example, in two dimensions, the coefficients of the $(m_1,m_2)$ and $(m_1,-m_2)$ directional terms must be equal in magnitude.

In contrast, the spectral path model introduced in this work assigns independent coefficients to each directional harmonic. Tensor-product Chebyshev expansions therefore correspond to a structured subspace of the directional model, defined by linear constraints that enforce coefficient tying across sign-related frequency vectors. By relaxing these constraints, the directional model defines a strictly larger hypothesis class. This increased flexibility allows the model to represent functions that cannot be expressed by a tensor-product Chebyshev expansion of comparable complexity, and in practice leads to improved approximation accuracy at fixed model size.

\subsection{Remarks on Approximation Guarantees}

The classical near-minimax optimality of Chebyshev polynomials is a statement about best uniform approximation in one dimension under degree-based truncation. These guarantees do not transfer directly to the setting considered here, for several reasons: the approximation problem is multivariate; coefficients are estimated from data rather than obtained by orthogonal projection; model complexity is controlled through sparse, data-dependent selection of frequency vectors; and optimisation is performed with respect to a squared-loss objective with regularisation rather than the uniform norm.

For these reasons, we do not claim minimax optimality of the resulting models. Instead, Chebyshev theory serves as a geometric and spectral grounding: the angular transformation and integer-frequency organisation provide a well-adapted coordinate system for smooth functions on bounded domains. Within this coordinate system, the directional basis offers a flexible and interpretable representation whose coefficients are chosen to minimise empirical error rather than to satisfy classical optimality criteria. Empirically, this relaxation yields compact models with strong predictive performance, as demonstrated in Section~\ref{sec:experiments}.

\section{Implementation Details of Greedy Spectral Path Selection}
\label{app:greedy-details}

This appendix provides additional details on the implementation of the greedy spectral path selection procedure described in Section~\ref{subsec:greedy-selection}.

\subsection{Candidate Generation}

We restrict attention to sparse frequency vectors
\[
\bm{m}\in\mathbb{Z}^D,
\qquad
\|\bm{m}\|_0 = k,
\qquad
k\in\mathcal{K},
\]
where $\mathcal{K}$ is a small set such as $\{1,2,3,4\}$. For fixed sparsity $k$, candidates are constructed by selecting supports of size $k$ and assigning positive integer coefficients whose sum defines the total order $L=\|\bm{m}\|_1$.

Candidates are enumerated in increasing order of $L$, so that low-frequency components are considered before more oscillatory ones.

Optionally, feature-importance ordering may be used to prioritise supports, based on simple univariate statistics computed on the transformed training data. This affects only the order of exploration, not the evaluation procedure.

\subsection{Primitive Rays and Harmonic Structure}

Each frequency vector $\bm{m}$ can be uniquely written as
\[
\bm{m} = r\,\bm{p},
\qquad
r\in\mathbb{Z}_{>0},
\qquad
\bm{p}\ \text{primitive}.
\]
In practice, we store primitive directions $\bm{p}$ together with harmonic orders $r$, so that features take the form
\[
\phi_{\bm{m}}(\bm{\theta}) = \cos\!\bigl(r\,\bm{p}^\top \bm{\theta}\bigr).
\]
This reduces redundant computation when multiple harmonics along the same direction are present.

\subsection{Streaming Normal Equation Updates}

Let $\mathcal{M}$ denote the current dictionary and $\Delta\mathcal{M}$ a candidate block. The augmented ridge system depends on
\[
\Phi_{\mathcal{M}}^\top \Phi_{\mathcal{M}},\quad
\Phi_{\mathcal{M}}^\top \Phi_{\Delta\mathcal{M}},\quad
\Phi_{\Delta\mathcal{M}}^\top \Phi_{\Delta\mathcal{M}},\quad
\Phi_{\mathcal{M}}^\top \bm{y},\quad
\Phi_{\Delta\mathcal{M}}^\top \bm{y}.
\]

These quantities are accumulated in a single pass over the training data, yielding block matrices
\[
G_{\mathrm{trial}}=
\begin{bmatrix}
G_{\mathrm{old}} & C \\
C^\top & G_{\mathrm{new}}
\end{bmatrix},
\qquad
b_{\mathrm{trial}}=
\begin{bmatrix}
b_{\mathrm{old}}\\
b_{\mathrm{new}}
\end{bmatrix}.
\]

This avoids materialising the full design matrix and allows efficient evaluation of each candidate block.

\subsection{Regularisation Strategy}

At the first iteration, all $\lambda\in\Lambda$ are evaluated and the best-performing value $\lambda_\star$ is selected. Subsequent iterations fix $\lambda_\star$ to reduce computational cost. After the greedy search terminates, an optional final resweep over $\Lambda$ may be performed.

\subsection{Adaptive Block Size and Early Stopping}

Paths are added in blocks of size $B$. The block size may be adjusted dynamically depending on validation improvement, subject to predefined bounds. Early stopping is triggered when validation performance fails to improve beyond a tolerance for several consecutive iterations.

\subsection{Summary}

These implementation choices enable efficient and stable construction of the spectral path dictionary while preserving exact evaluation of candidate contributions.

\section{Runtime and Wall-Clock Performance}
\label{app:implementation}
\begin{table}[H]
\centering
\caption{Total experiment time in seconds across datasets. Best result per dataset in bold.}
\label{tab:results}
\begin{tabular}{l l c c c c c}
\toprule
Source & Dataset              & Spectral Paths & Ridge &  MLP  & XGBoost \\
\midrule
UCI    & Concrete             & 1.0 & $\mathbf{3.0\times 10^{-3}}$
 &  28 & 0.53 \\
UCI    & Energy (Heating)     & 0.4   & $\mathbf{2.9\times 10^{-3}}$  & 28  & 0.27  \\
UCI    & Energy (Cooling)     & 0.4   & $\mathbf{7.5\times 10^{-3}}$  & 26 & 0.57  \\
UCI    & Superconductivity    & 6.3   & $\mathbf{2.2\times10^{-2}}$  & 120  & 3.7  \\
UCI    & Wine Quality         & 4.4    & $\mathbf{7.2\times 10^{-3}}$ & 16  & 3.4  \\
UCI    & Phishing Websites    & 5.5  & $\mathbf{7.1\times10^{-3}}$  & 34  & 0.37 \\
OpenML & Concrete Slump       & 0.4 & $\mathbf{1.5\times10^{-2}}$ & 8.5 & 3.5$\times10^{-2}$ \\
OpenML & Yacht Hydrodynamics  & 0.1 & $\mathbf{2.8\times10^{-3}}$ & 32 & 4.1$\times10^{-2}$ \\
OpenML & Cancer Drug Response & 1.4 & $\mathbf{1.8\times10^{-2}}$ & 11 & 9.7 \\
OpenML & Aquatic Toxicity     & 0.2 & $\mathbf{1.8\times10^{-3}}$ & 6.9 & 0.12 \\
OpenML & Izmir Weather        & 0.2 & $\mathbf{2.1\times 10^{-3}}$ & 6.9 & 0.11 \\
OpenML & Ankara Weather       & 0.1 & $\mathbf{1.8\times 10^{-3}}$ & 4.1 & 9.9$\times10^{-2}$ \\
PMLB   & Echocardiogram       & 1.5 & $\mathbf{1.7\times 10^{-3}}$ & 23.5 & 0.63 \\
PMLB   & Wind Speed           & 7.3 & $\mathbf{3.3\times 10^{-3}}$ & 14 & 0.75 \\
PMLB   & CPU Utilisation      & 5.8 & $\mathbf{3.1\times10^{-2}}$ & 26 & 0.52 \\
\bottomrule
\end{tabular}
\end{table}
Timings are wall-clock totals averaged over 5 runs; standard deviations were below 10\% of the mean for all methods and are omitted for clarity.
\end{document}